\documentclass[runningheads]{llncs}
\usepackage{algorithm}
\usepackage{algpseudocode}
\usepackage{caption}
\usepackage{amsmath}
\usepackage{subcaption}
\usepackage{comment}
\usepackage{graphicx}
\usepackage{hyperref}
\usepackage{colortbl}
\begin{document}
\title{Search-time Efficient Device Constraints-Aware Neural Architecture Search}
%
%
\author{Anonymous Author}
\author{Oshin Dutta \and
Tanu Kanvar \and
Sumeet Agarwal}
%
%
\institute{Indian Institute of Technology \\
\email{\{oshin.dutta,sumeet\}@ee.iitd.ac.in, kanvar.tanu@gmail.com}}
%
\maketitle              
\begin{abstract}
Edge computing aims to enable edge devices, such as IoT devices, to process data locally instead of relying on the cloud. However, deep learning techniques like computer vision and natural language processing can be computationally expensive and memory-intensive. Creating manual architectures specialized for each device is infeasible due to their varying memory and computational constraints. To address these concerns, we automate the construction of task-specific deep learning architectures optimized for device constraints through Neural Architecture Search (NAS). We present DCA-NAS, a principled method of fast neural network architecture search that incorporates edge-device constraints such as model size and floating-point operations. It incorporates weight sharing and channel bottleneck techniques to speed up the search time. Based on our experiments, we see that DCA-NAS outperforms manual architectures for similar sized models and is comparable to popular mobile architectures on various image classification datasets like CIFAR-10, CIFAR-100, and Imagenet-1k. Experiments with search spaces---DARTS and NAS-Bench-201 show the generalization capabilities of DCA-NAS. On further evaluating our approach on Hardware-NAS-Bench, device-specific architectures with low inference latency and state-of-the-art performance were discovered.
\end{abstract}

\section{Introduction}
In recent years, there has been significant progress in developing Deep Neural Network (DNN) architectures~\cite{sandler2018mobilenetv2,zhang2018shufflenet,srivastava2021variational} for edge and mobile devices.However, designing DNN architectures for specific hardware constraints and tasks is a time-consuming and computationally expensive process~\cite{cai_once-for-all_2020}. To address this, Neural Architecture Search (NAS) ~\cite{baker2017accelerating,refbib1,zhou2019bayesnas} has become popular as it discovers optimal architectures given a task and network operations. Despite its success, traditional NAS techniques cannot guarantee optimal architecture for specific devices with hardware constraints such as storage memory and maximum supported FLOPs.
To address this concern, researchers have developed hardware-aware algorithms~\cite{tan2019platform,cai2019proxylessnas} that find optimal device architectures with low resource training overhead and search time. These methods often use inference latency~\cite{cai2019proxylessnas}, FLOPs~\cite{tan2019platform} or a combination of hardware metrics~\cite{tan2019platform} as constraints scaled by a tunable factor. However, the time to tune the scaling factor is often not considered within the NAS search time and can be ten times the reported search time.
To address these issues, we propose the Device Constraints-Aware NAS (DCA-NAS), a principled differentiable NAS method that introduces total allowable model size or floating-point operations (FLOPs) as constraints within the optimization problem, with minimal hyper-parameter tuning. Unlike inference latency  which is task dependent, FLOPs and memory are specified with a given hardware and thus are appropriate for our generic method. The approach is adaptable to other hardware metrics such as energy consumption or inference latency using additional metric-measuring functions.
The paper make the following significant contributions:
\begin{itemize}
\item It introduces a fast method that uses weight sharing among operations in the search space and channel bottleneck, along with a differentiable resource constraint, for continuous exploration of the search space.
\item A training pipeline that allows a user to input device memory or FLOPs and search for optimal architecture with minimal hyper-parameter tuning.
\item Our extensive experimentation on vision datasets- CIFAR-10, CIFAR-100, TinyImagenet, Imagenet-1k and inference-latency comparisons of trained models on Hardware-NAS-bench  demonstrate the efficiency of our method. The generalization of our method to different search spaces is shown with experiments on DARTS and NAS-Bench.

\end{itemize}
\begin{figure}[t]
\centering
\includegraphics[width=\linewidth]{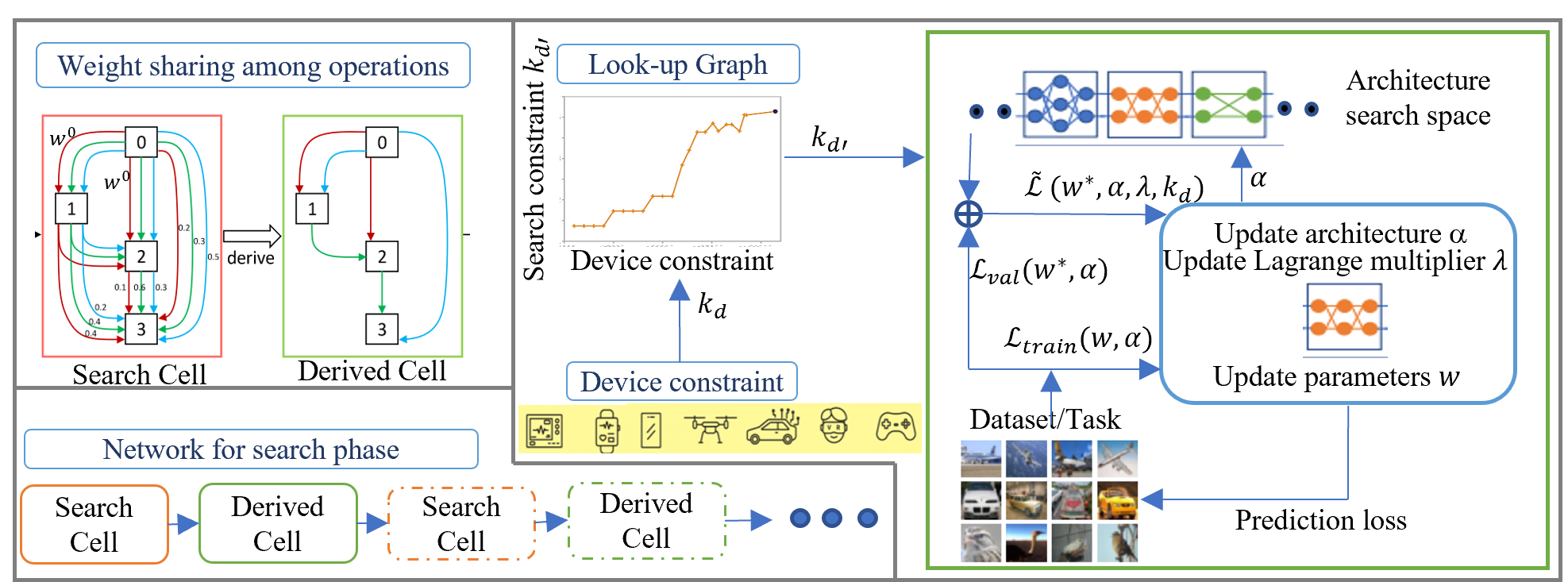}
    \caption{\small DCA-NAS framework:Weight sharing in the search space and Derived cells lowers the search time from other DNAS. Target device constraint is used to query search constraint from look-up graph for constrained optimization.}
    \label{figframework}
    \vspace{-6mm}
\end{figure}
 \vspace{-4mm}
\section{Related Work}
\textbf{Neural Architecture Search}
Popular approaches~\cite{han2017deep,kim2020puzzle,abai2020densenet} designed architectures for high performance on specific tasks or datasets with the traditional deep learning perspective that bigger is better, resulting in computationally and memory-intensive inference on edge devices. Network pruning~\cite{han2016deep}, channels removal~\cite{liu2017learning,srivastava2021variational} and weights/activations quantization~\cite{courbariaux2016binaryconnect,zhu2017trained} can compress architectures, but require pre-training, hyperparameter tuning, and often lack transferability.Neural Architecture Search (NAS) methods such as Reinforcement Learning~\cite{Pham2018EfficientNA,cai2019proxylessnas}, Evolutionary Learning~\cite{Elsken2019EfficientMN,refbib2} and Differentiable Neural Architecture Search (DNAS)~\cite{liu2018darts,xie2018snas} can automatically search for architectures without user intervention, and can transfer across similar tasks. DNAS with surrogate metrics~\cite{xiao_shapley-nas_2022,zheng_neural_2022} have also been used to explore the architecture search space. However, architectures found by DNAS methods are not optimized for deployment on edge devices and smaller models obtained by reducing layers or channels are often sub-optimal.\\
\textbf{Hardware-aware Neural Architecture search}
Certain NAS methods optimize~\cite{cai2019proxylessnas,wu2019fbnet,cai_once-for-all_2020,jiang_eh-dnas_2021} for constraints such as latency, inference speed~\cite{wu_compiler-aware_2022}, FLOPS~\cite{tan2019platform,tan2019efficientnet}, memory usage~\cite{lin2020mcunet}. Some use a separate DNN to predict constraint metrics and evolutionary search to obtain hardware-aware optimal models~\cite{tan2019platform,cai_once-for-all_2020}, while others consider real-time latencies of edge devices or provide specific architectures for specific devices~\cite{lyu_resource-constrained_2022,chu2021discovering}. However, these methods require significant search time and tuning of scaling factors controlling the trade-off between the performance and the constraint, and do not always account for optimal architectures. In contrast, we use a differentiable hardware-aware objective function with generic hardware metrics, and do not require a tunable scaling factor.  
 Certain methods~\cite{cai_once-for-all_2020,munoz_enabling_2021,ding2022nap} train a supernet first and then search for a smaller architecture, but this is only efficient when there are more than fifteen different edge devices with different limitations or deployment scenarios ~\cite{cai_once-for-all_2020} as training the supernet takes huge resources-32 V100s taking about 1,200 GPU hours. Search stage followed by evaluation, as done in our approach is more efficient when the different number of possible edge devices is less than fifteen.
\vspace{-3mm}
\section{DCA-NAS: Device Constraints Aware Fast Neural Architecture Search}
We present the preliminary gradient-based NAS objective function in section~\ref{prelim} and then formulate the problem of incorporating the hardware-awareness in NAS as a constrained optimization problem in section~\ref{cons} followed by techniques to reduce the search time in section~\ref{searchtime}. The framework of our approach is illustrated in Figure~\ref{figframework}.
\vspace{-3mm}
\subsection{Gradient-based NAS Objective Function} 
\vspace{-2mm}
\label{prelim}Popular DNAS techniques~\cite{liu2018darts,yang2021towards} have two stages, the search phase and the evaluation phase. During the search phase, given a task or a dataset the techniques search for a network of cells, which are directed acyclic graphs with $N$ nodes. The edges of the graph are network layers, whose operations are to be selected from a pre-defined set $\mathcal{O}$ containing operations such as 3x3 separable convolution and identity operations with trainable weights $w_{o}$.
The search is made differentiable by making the choice of a particular operation to be a softmax of architecture weights $\alpha$ of all operations. Thus, the intermediate output $z_j$ at node $j$ is given by, 
\begin{equation}
\small
z_{j}=\sum_{o \in \mathcal{O}} \frac{\exp \left\{\alpha_{o}^{i, j}\right\}}{\sum_{o^{\prime} \in \mathcal{O}} \exp \left\{\alpha_{o^{\prime}}^{i, j}\right\}} \cdot o\left(w_{o}^{i,j},\mathbf{z}_{i}\right)
\label{eq0}
\end{equation}
\subsection{DCA-NAS formulation}
\label{cons}
Previous DNAS approaches~\cite{liu2018darts,xu2019pc,yang2021towards} did not focus on searching architectures specifically for inference on resource-constrained devices. In contrast, we formulate the DNAS objective function as a constrained optimization problem by incorporating device resource constraints (memory or FLOPs) in the search objective function. The constrained bi-level optimization problem is written as,
\begin{equation}
\label{eq2}
\small
\begin{array}{ll}\min _{\alpha} & \mathcal{L}_{\text {val }}\left(w^{*}(\alpha), \alpha\right)
\\ \text { s.t. } & w^{*}(\alpha)=\operatorname{argmin}_{w} \mathcal{L}_{\text {train }}(w, \alpha)
\\ \text { s.t. } & k_{s}(\alpha) \leq K_{d}
\end{array} 
\end{equation} 
where training dataset is split into $train$ and $val$ to optimize $w$ and $\alpha$ simultaneously in each iteration subject to the constraint that the architecture's number of parameters or FLOPs $k_{s}$ must be less than or equal to the device resource constraint $K_{d}$. The following equation calculates the architecture's number of parameters or FLOPs during search given the number of cells$c_{n}$ . Our method can also be adapted to use other metrics such as latency and energy consumption with additional metric measuring functions.

\begin{equation}
\vspace{-3mm}
\small
\label{eq4}
 k_{s}(\alpha)= c_{n} \sum_{(i,j)\in N} \sum_{o \in \mathcal{O}} \frac{\exp\{{\alpha_{o}^{i, j}\}} * b\left(o\right)}{\sum_{o^{\prime} \in \mathcal{O}} \exp\{{\alpha_{o^{\prime}}^{i, j}}\}}   
\end{equation}

\subsubsection{Tackling the difference in search and evaluation networks}
The size of the architecture in the search phase $k_{s}$ is different from the architecture size in evaluation phase due to the softmax weighting factor in equation~\ref{eq4} (demonstration can be found in the appendix). To address this, we introduce a tighter bound on the search constraint $K_{d^{\prime}}$, which is less than the device resource constraint $K_d$. A lookup graph (LUG) needs to be made for each dataset by varying $K_{d^\prime}$ within appropriate bounds and running the algorithm until convergence each time to obtain the corresponding device resource constraint $K_d$. The computation time of the LUG can be reduced by running the searches in parallel. Thus, on incorporating the tighter constraint by looking-up the graph for the given device resource constraint $K_d$ along with the trainable Lagrange multiplier $\lambda$ in Equation~\ref{eq2}, the objective function is re-written as, 
\begin{equation}
\label{eq3a}
\small
\begin{array}{c}
  \widetilde{\mathcal{L}} =\mathcal{L}_{\text {val }}\left(w^{*}(\alpha), \alpha\right)
   +\lambda (k_{s}(\alpha)-LUG(K_{d}))
    \\ \text { s.t. } w^{*}(\alpha)=\operatorname{argmin}_{w} \mathcal{L}_{\text {train }}(w, \alpha)
\end{array}
\end{equation}
\begin{table}[t]
\vspace{-4mm}
\centering
\caption{\small Performance comparison of architectures evaluated on visual datasets- CIFAR-10 and TinyImagenet. '(CIFAR-10)' indicates search with CIFAR-10. 'X M' in 'DCA-NAS-X M' denotes the input memory constraint. RCAS- Resource Constrained Architecture Search }
\resizebox{0.8\textwidth}{!}{
\begin{tabular}{lcccccc}
\hline
\textbf{Dataset} & \textbf{Search}&\textbf{Method} & \textbf{Accuracy} & \textbf{Parameters} & \textbf{GPU}\\
&\textbf{Strategy}&&(\%)&(Million)&Hours \\
\hline

CIFAR-10 &manual&PyramidNet-110 (2017) \cite{han2017deep} &95.74 &3.8&-\\
&manual&VGG-16 pruned (2017) \cite{he2017channel}& 93.4 &5.4&-\\

&evolution&Evolution + Cutout (2019)~\cite{abc}&96.43&5.8&12\\

&random&NAO Random-WS (2019)~\cite{real2019regularized}&96.08&3.9&7.2\\
&gradient&ENAS + micro + Cutout (2018)~\cite{Pham2018EfficientNA}&96.46&4.6&12\\
&gradient&DARTS + Cutout (2nd) (2018)~\cite{liu2018darts}&97.24$\pm$0.09&3.3&24\\
&gradient&SNAS + Cutout (2018)~\cite{xie2018snas}&97.15&2.8&36\\
&gradient&PC-DARTS (2019)~\cite{xu2019pc}&97.43$\pm$ 0.07&3.6&2.4\\
&gradient&SGAS (2020)~\cite{li2020sgas}&97.34&3.7&6\\
&gradient&DrNAS (2020)~\cite{chen2020drnas}& 97.46 ± 0.03 &4.0 &9.6 \\
&gradient&DARTS+PT (2021)~\cite{wang2021rethinking} & 97.39 ± 0.08 &3.0 &19.2\\
&gradient&Shapley-NAS (2022)~\cite{xiao_shapley-nas_2022}& 97.53 ± 0.04 &3.4& 7.2 \\
\rowcolor[gray]{0.8}
&RCAS&DCA-NAS- 3.5 M (CIFAR-10)&97.2$\pm$0.09&\textbf{3.4}&\textbf{1.37}\\
\hline

Tiny ImageNet&manual&SqueezeNet (2016)~\cite{iandola2016squeezenet}&54.40&-&-\\
&manual&PreActResNet18 (2020)~\cite{kim2020puzzle} &63.48 & - & - \\
&manual&ResNet18 (2016)~\cite{he2016deep} &58.4 & 6.4 & -  \\
&manual& DenseNet (2020)~\cite{abai2020densenet} &62.73 & 11.8 & - \\
&gradient&DARTS+ Cutout (2018)~\cite{liu2018darts}&62.15$\pm$0.15&7.3&219\\
\rowcolor[gray]{0.8}
&RCAS&DCA-NAS- 3.5 M& 61.34$\pm$0.09&\textbf{3.5}&\textbf{12.5}\\
\rowcolor[gray]{0.8}
&RCAS&DCA-NAS- 3.5 M (CIFAR-10)&61.4$\pm$0.15&\textbf{3.4}&\textbf{1.37}\\
\hline
\end{tabular}
}
\label{tab1}
\vspace{-3mm}
\end{table}
\begin{figure}[t]
\centering
\includegraphics[width=0.95\linewidth]{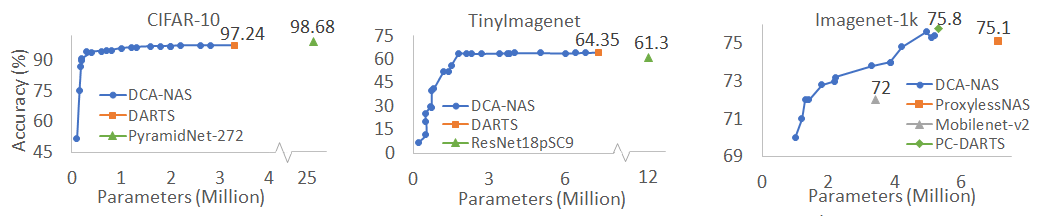}
    \caption{\small Plots show that DCA-NAS method discovers models with fewer parameters than other NAS methods and manual architectures without sacrificing prediction performance to a large extent.}
    \label{figTrade}
    \vspace{-5mm}
\end{figure}
\vspace{-5mm}
\subsection{Techniques to reduce search time}
\label{searchtime}
\textbf{Channel Bottleneck} We use convolutional layers of 1x1 kernel to reduce the depth of output channels of operations in the search space to save computation time and memory overhead.\\
\textbf{Derived Cell and Weight sharing}. During architecture search, only one cell with trainable $\alpha$ is used to optimize architecture parameters. The target network for inference is built by stacking cells with architectures derived from highly weighted operations. This can be done during search by deriving the other cell architectures from the first at each iteration~\cite{yang2021towards}. The arrangement of the cells for search is given in the appendix. This derived cell saves computation and memory overhead. A weight sharing strategy~\cite{yang2021towards} among same operations with the same originating node $i$ to all nodes $i<j<N$ has been applied within a cell. This is motivated by the observation that non-parametric operations operating on the representation of a node produce the same feature map irrespective of the output node and thereby extended to parametric operations. Thus, Equation~\ref{eq0} may be re-written to the following,
\begin{equation}
\small
z_{j}=\sum_{o \in \mathcal{O}} \frac{\exp \left\{\alpha_{o}^{i, j} \right\}}{\sum_{o^{\prime} \in \mathcal{O}} \exp \left\{\alpha_{o^{\prime}}^{i, j} \right\}} \cdot o\left(w_{o}^{i},\mathbf{z}_{i}\right)
\label{eq7}
\end{equation}
\vspace{-1mm}
\section{Experimental Results}
\vspace{-1mm}
Our approach is evaluated on two search spaces- DARTS and NAS-Bench with vision datasets- CIFAR10, TinyImagenet, Imagenet-16-20 and Imagenet-1k. The details of the search space and implementation is given in the appendix
\subsection{Results on DARTS search space}
\begin{table}[t]
\vspace{-4mm}
\caption{\small Performance and comparison of architectures evaluated on Imagenet-1k. The label "(Imagenet)" indicates that the architecture has been searched and evaluated on Imagenet-1k.; else it is searched on CIFAR-10. 'X M' in 'DCA-NAS-X M' denotes the input memory constraint}
\label{tab2}
\centering
\resizebox{0.85\textwidth}{!}{
\begin{tabular}{lcccccc}
\hline
\textbf{Method} & \multicolumn{2}{c}{\textbf{Test Error (\%)}} & \textbf{Parameters}  & \textbf{FLOPS} & \textbf{Search Cost}&\textbf{Search}\\
&\textbf{top-1}&\textbf{ top-5}& \textbf{(Mil)} &\textbf{(Mil)}& \textbf{(GPU days)}&\textbf{Strategy}\\
\hline
Inception-v1 (2015)~\cite{szegedy2015going} & 30.2 &10.1& 6.6& 1448& -& manual \\
MobileNet\_V1 (2017)~\cite{howard2017mobilenets}& 29.4& 10.5& 4.2 &569& -& manual\\
MobileNet\_V2 (2018)~\cite{sandler2018mobilenetv2}& 72.0 &91.0 &3.4& 300&-& manual\\
ShuffleNet 2× (v2) (2018)~\cite{ma2018shufflenet} & 25.1& -& 5 &591 &- &manual\\
\hline
MnasNet-92 (2020)~\cite{he2020milenas}&25.2& 8.0 &4.4 &388 & -& RL\\
AmoebaNet-C (2019)~\cite{real2019regularized} & 24.3 &7.6& 6.4 &570& 3150 &evolution\\
\hline
DARTS+Cutout (2018)~\cite{liu2018darts}& 26.7 &8.7 &4.7& 574 &1.0& gradient\\
SNAS (2018)~\cite{xie2018snas} & 27.3& 9.2& 4.3& 522& 1.5& gradient\\
GDAS (2019)~\cite{dong2019searching} & 26.0 &8.5&5.3& 545& 0.3&gradient \\
BayesNAS (2019)~\cite{zhou2019bayesnas} &26.5&8.9& 3.9 &-& 0.2& gradient\\
P-DARTS (2018)~\cite{Pham2018EfficientNA}& 24.4& 7.4& 4.9& 557 &0.3& gradient\\
SGAS (Cri 1. best) (2020)~\cite{li2020sgas} &\textbf{24.2 }&\textbf{7.2}&5.3 &585& 0.25 &gradient\\
SDARTS-ADV (2020)~\cite{chen2020stabilizing} &25.2&7.8& 6.1 &-& 0.4&gradient\\
Shapley-NAS (2022)~\cite{xiao_shapley-nas_2022} &24.3 &-& 5.1 &566& 0.3&gradient\\
\hline
RC-DARTS (2019)~\cite{jin2019rc} & 25.1 & 7.8 & 4.9 &590 & 1 & RCAS\\
\rowcolor[gray]{0.8}
DCA-NAS& 25.1& 8.1& \textbf{5.1}& 578&\textbf{ 0.06}& RCAS\\
\hline

ProxylessNAS (GPU) (2019)~\cite{cai2019proxylessnas}(Imagenet)& 24.9& 7.5 &7.1 &465& 8.3& gradient\\
PC-DARTS (2019)~\cite{xu2019pc} (Imagenet)& 24.2& 7.3& 5.3& 597& 3.8& gradient\\
DrNAS (2020)~\cite{chen2020drnas} (Imagenet)  &24.2 &7.3&5.2 &644 &3.9&gradient\\
DARTS+PT (2021)~\cite{wang2021rethinking} (Imagenet) & 25.5 &-&4.7& 538 &3.4&gradient\\
Shapley-NAS (2022)~\cite{xiao_shapley-nas_2022} (Imagenet) &23.9 &-&5.4 &582 &4.2&gradient\\
\hline
RCNet-B (2019)~\cite{xiong2019resource} (ImageNet) &25.3& 8.0 &4.7& 471&  9 &RCAS\\
\rowcolor[gray]{0.8}
DCA-NAS- 5.5 M(Imagenet) &24.4& 7.2&5.3& 597& \textbf{1.9}& RCAS\\
\hline
\end{tabular}}
\vspace{-3mm}
\end{table}
\vspace{-1mm}
\subsubsection{Transferability- learning of coarse features during search.} We transfer the architecture searched on CIFAR-10 to train and evaluate the model weights on TinyImagenet in Table~\ref{tab1} and ImageNet-1k in Table~\ref{tab2}. This transferred model yields higher performance than manually designed architectures~\cite{sandler2018mobilenetv2,ma2018shufflenet} for the target dataset. It is observed that performance of the transferred model is comparable to the architecture searched on the target dataset itself which can be attributed to the architecture learning coarse features than objects during search.
\vspace{-5mm}
\subsubsection{Performance versus Device-Constraints trade-off}
DCA-NAS discovers  2 to 4\% better-performing architectures than manual designs with a memory constraint of 3.5 million parameters on CIFAR-10 and similar performance on TinyImagenet as in Table~\ref{tab1}.
On Imagenet-1k, DCA-NAS yields models with similar performance to other NAS methods~\cite{xiao_shapley-nas_2022,chen2020drnas,xu2019pc} with a constraint of 5.5 million parameters (taken to yield similar sized models as other NAS methods) as in Table~\ref{tab2}. We vary the input device resource constraint and plot the performance of the searched models against the number of parameters in Figure \ref{figTrade}. As observed, DCA-NAS searched models can yield 15x lower sized models than manual architectures like PyramidNet-272~\cite{han2017deep} with at most 1\% reduction in accuracy on CIFAR-10. On TinyImagenet, DCA-NAS yields models similar in performance but 6x smaller in size than the manual Resnet variant. In comparison to ProxylessNAS~\cite{cai2019proxylessnas} for Imagenet-1k, DCA-NAS yields 32\% smaller model in terms of model parameters for similar accuracy. In comparison to DNAS methods~\cite{liu2018darts,xu2019pc} for each of the three datasets, we observe that the performance of the DCA-NAS searched models is retained to a certain extent as resources are further limited after which the model performance degrades. DCA-NAS model of similar size has the advantage of better performance (by 1\%) and being automatically searched over MobileNet-v2~\cite{sandler2018mobilenetv2}, a manually designed network on Imagenet-1k. 
\vspace{-6mm}
\subsubsection{Search time comparison}
For evaluation on TinyImagenet in Table~\ref{tab1}, the architecture searched on CIFAR-10 with DCA-NAS yields model in the lowest search time which indicates the search-time efficiency of the transferability property. Our method requires about 4x lower search cost than SGAS~\cite{li2020sgas} which performs the best among the other transferred architectures and 16x lower search time than the other resource-constrained approach~\cite{jin2019rc} for similar performance as seen in Table~\ref{tab2}. Moreover, ProxylessNAS~\cite{cai2019proxylessnas} takes about 4x more search time than DCA-NAS whereas PC-DARTS takes about 2x more search time with no capability to constraint model size.
\vspace{-3mm}
\begin{figure}[t]
\begin{center}
\includegraphics[width=1\linewidth]{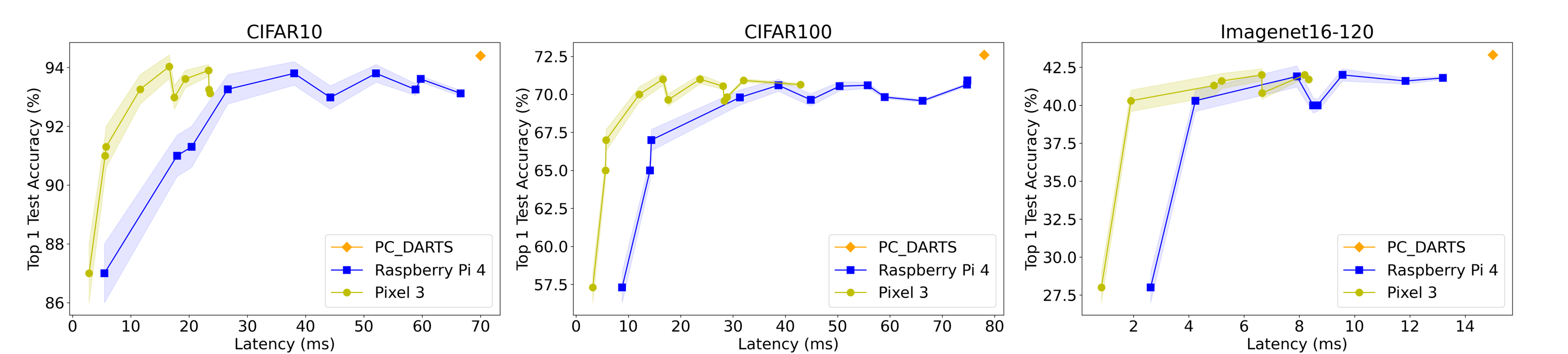}
\end{center}
\vspace{-3mm}
    \caption{\small Plots show DCA-NAS searched models with similar performance but lower inference latency (on two devices- Pixel 3 and Raspberry Pi 4) to previous SOTA NAS method- PC-DARTS when evaluated on NAS-Bench dataset.}
    \label{figLat}
    \vspace{-5mm}
\end{figure}
\subsection{Results on NAS-Bench-201 search space}
\subsubsection{Performance and Latency comparisons on different devices}
Our method reports the mean by averaging over five runs with different random seed. Figure \ref{figLat} compares the performance of models searched with DCA-NAS and PC-DARTS by varying the latency constraints. It shows that unlike PC-DARTS, DCA-NAS can search for more efficient models which have lower inference latency for similar test accuracy. Moreover, we observe that models with similar performance have lower latency when tested on Pixel 3 than on Raspberry Pi 4 due to a faster RAM in Pixel 3. 
DCA-NAS takes the lowest search time among all the NAS methods due to the addition of search-time-efficient techniques while being at-par in terms of performance across all datasets.
\vspace{-3mm}
\begin{figure}[t]
\centering
    \begin{subfigure}[b]{0.48\textwidth}
         \centering
         \includegraphics[width=\textwidth]{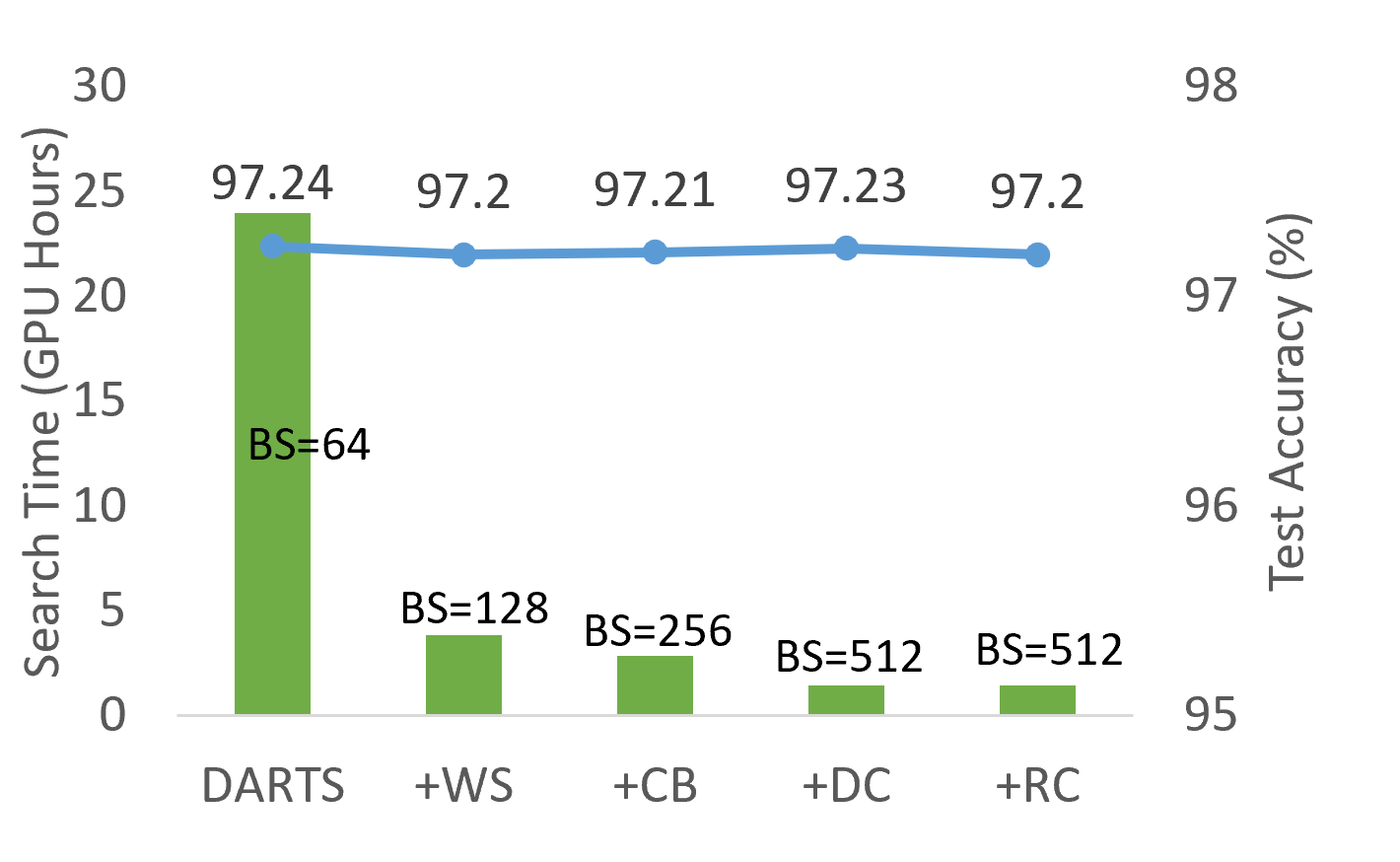}
         \caption{}
         \label{ablation}
         
     \end{subfigure}
    \begin{subfigure}[b]{0.48\textwidth}
         \centering
         \includegraphics[width=0.9\textwidth]{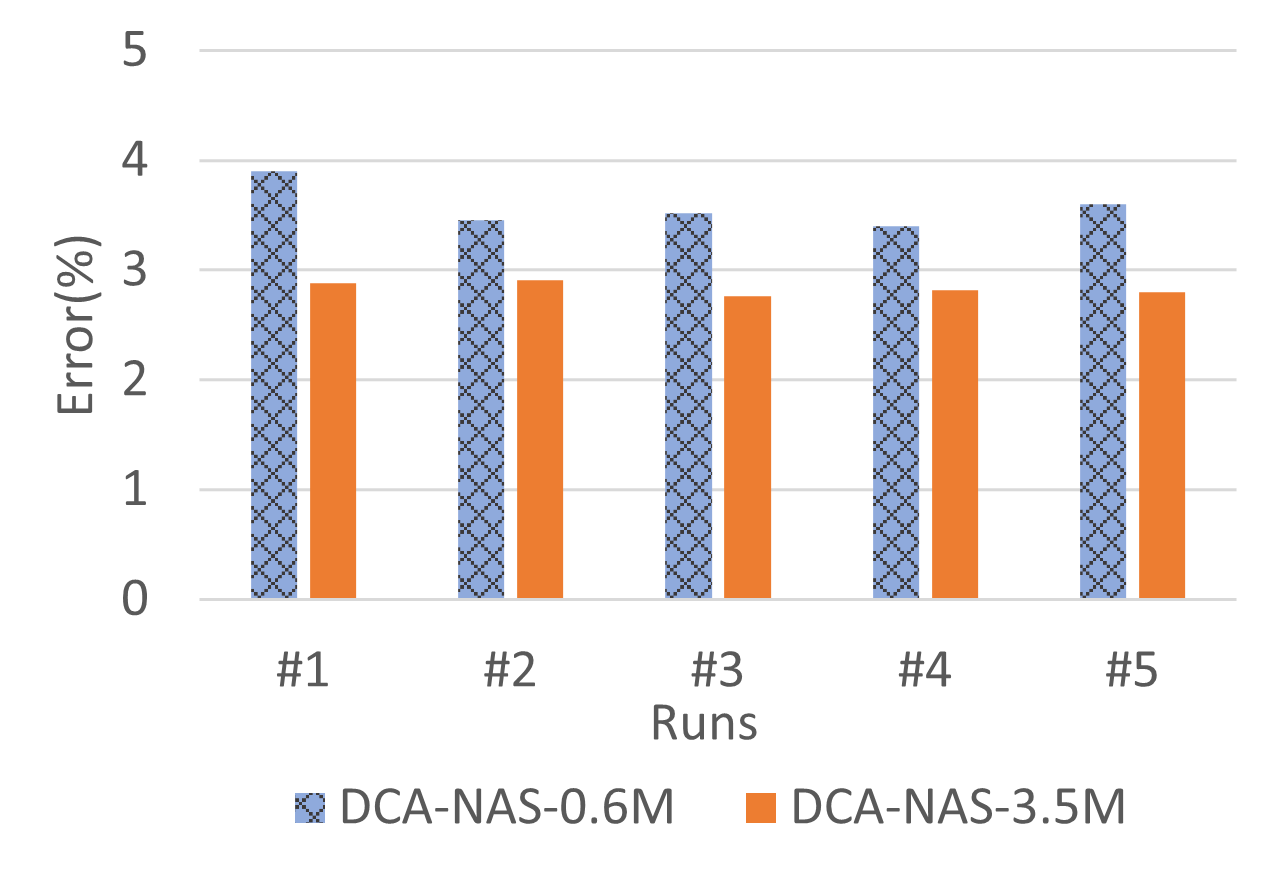}
         \caption{}
         \label{stability}
    \end{subfigure}
    \vspace{-4mm}
    \caption{\small (a) Ablation study with CIFAR-10 dataset- Each component added to DARTS leads to the reduction in the search cost of DCA-NAS while performance is retained. WS- Weight Sharing, CB- Channel Bottleneck, DC- Derived Cell, RC- Resource Constraint, BS- Batch Size (b) Shows stability of performance of DCA-NAS searched models for runs with varying seeds on CIFAR-10 dataset.}
\vspace{-4mm}
\end{figure}
\section{Ablation Study}
\vspace{-3mm}
\textbf{Effectiveness of various algorithmic augmentations for faster search:} We analyze the effectiveness of algorithmic augmentations mentioned preciously~\ref{searchtime} to reduce search cost in our study. We sequentially add weight sharing, channel bottleneck, and derived cells to the baseline DARTS~\cite{liu2018darts} method and measure search time and accuracy. Weight sharing, channel bottleneck, and derived cells was observed to significantly reduce search memory overhead, enabling us to use larger batch sizes and reducing overall search cost as seen in Figure ~\ref{ablation}. Adding the resource-constraint in the final DCA-NAS method negligibly increases search cost while maintaining performance.   \\
\textbf{Stability of the approach:}
We test stability by running the search algorithm independently five times with different initial seeds and the same constraints and hyperparameters. The architectures found during each run have similar performance when re-trained and evaluated as shown in Fig.~\ref{stability}. Smaller models have lower performance due to restrictions in model complexity compared to larger models.
\vspace{-3mm}
\section{Conclusion}
\vspace{-3mm}
We present DCA-NAS, a device constraints-aware neural architecture search framework which discovers architectures optimized to the memory and computational constraints of an edge device in a time-efficient manner. It does so by incorporating a constraint in terms of the number of parameters or floating point operations (FLOPs) in the objective function with the help of a Lagrange multiplier. DCA-NAS in essence searches for a Pareto optimal solution given the edge device memory or FLOPs constraint. Moreover, it enables architecture search with search cost 4 to 17 times lower than the previous state-of-the-art Hardware-aware NAS approaches. DCA-NAS can discover models with size about 10 to 15 times lower than manually designed architectures for similar performance. In comparison to DARTS and its other NAS variants, DCA-NAS can discover models upto 3x smaller in size with similar performance. This hardware-aware approach can be generalized to any future updates to differential neural architecture search and possibly to training-free methods of NAS with some adaptation.

\section*{Acknowledgement}
We thank the anonymous reviewers; Profs. Surendra Prasad and Brejesh Lall of IIT Delhi; and colleagues at Cadence India for their valuable feedback and inputs. This research is supported by funding from Cadence India; the first author is also supported by a fellowship from the Ministry of Education, India.
%
%
\bibliographystyle{splncs04}
\begin{small}
\bibliography{egbib}

\begin{thebibliography}{10}
\providecommand{\url}[1]{\texttt{#1}}
\providecommand{\urlprefix}{URL }
\providecommand{\doi}[1]{https://doi.org/#1}

\bibitem{abai2020densenet}
Abai, Z., Rajmalwar, N.: Densenet models for tiny imagenet classification
  (2020)

\bibitem{baker2017accelerating}
Baker, B., Gupta, O., Raskar, R., Naik, N.: Accelerating neural architecture
  search using performance prediction (2017)

\bibitem{cai_once-for-all_2020}
Cai, H., Gan, C., Wang, T., Zhang, Z., Han, S.: Once-for-{All}: {Train} {One}
  {Network} and {Specialize} it for {Efficient} {Deployment} (Apr 2020),
  \url{http://arxiv.org/abs/1908.09791}, arXiv:1908.09791 [cs, stat]

\bibitem{cai2019proxylessnas}
Cai, H., Zhu, L., Han, S.: Proxylessnas: Direct neural architecture search on
  target task and hardware (2019)

\bibitem{chen2020stabilizing}
Chen, X., Hsieh, C.J.: Stabilizing differentiable architecture search via
  perturbation-based regularization. In: International conference on machine
  learning. pp. 1554--1565. PMLR (2020)

\bibitem{chen2020drnas}
Chen, X., Wang, R., Cheng, M., Tang, X., Hsieh, C.J.: Drnas: Dirichlet neural
  architecture search. arXiv preprint arXiv:2006.10355  (2020)

\bibitem{chu2021discovering}
Chu, G., Arikan, O., Bender, G., Wang, W., Brighton, A., Kindermans, P.J., Liu,
  H., Akin, B., Gupta, S., Howard, A.: Discovering multi-hardware mobile models
  via architecture search. In: Proceedings of the IEEE/CVF Conference on
  Computer Vision and Pattern Recognition. pp. 3022--3031 (2021)

\bibitem{courbariaux2016binaryconnect}
Courbariaux, M., Bengio, Y., David, J.P.: Binaryconnect: Training deep neural
  networks with binary weights during propagations (2016)

\bibitem{ding2022nap}
Ding, Y., Wu, Y., Huang, C., Tang, S., Wu, F., Yang, Y., Zhu, W., Zhuang, Y.:
  Nap: Neural architecture search with pruning. Neurocomputing  \textbf{477},
  85--95 (2022)

\bibitem{dong2019searching}
Dong, X., Yang, Y.: Searching for a robust neural architecture in four gpu
  hours. In: Proceedings of the IEEE/CVF Conference on Computer Vision and
  Pattern Recognition. pp. 1761--1770 (2019)

\bibitem{Elsken2019EfficientMN}
Elsken, T., Metzen, J.H., Hutter, F.: Efficient multi-objective neural
  architecture search via lamarckian evolution. arXiv: Machine Learning  (2019)

\bibitem{han2017deep}
Han, D., Kim, J., Kim, J.: Deep pyramidal residual networks. In: Proceedings of
  the IEEE conference on computer vision and pattern recognition. pp.
  5927--5935 (2017)

\bibitem{han2016deep}
Han, S., Mao, H., Dally, W.J.: Deep compression: Compressing deep neural
  networks with pruning, trained quantization and huffman coding (2016)

\bibitem{he2020milenas}
He, C., Ye, H., Shen, L., Zhang, T.: Milenas: Efficient neural architecture
  search via mixed-level reformulation. In: Proceedings of the IEEE/CVF
  Conference on Computer Vision and Pattern Recognition. pp. 11993--12002
  (2020)

\bibitem{he2016deep}
He, K., Zhang, X., Ren, S., Sun, J.: Deep residual learning for image
  recognition. In: Proceedings of the IEEE conference on computer vision and
  pattern recognition. pp. 770--778 (2016)

\bibitem{he2017channel}
He, Y., Zhang, X., Sun, J.: Channel pruning for accelerating very deep neural
  networks. In: Proceedings of the IEEE International Conference on Computer
  Vision. pp. 1389--1397 (2017)

\bibitem{howard2017mobilenets}
Howard, A.G., Zhu, M., Chen, B., Kalenichenko, D., Wang, W., Weyand, T.,
  Andreetto, M., Adam, H.: Mobilenets: Efficient convolutional neural networks
  for mobile vision applications (2017)

\bibitem{iandola2016squeezenet}
Iandola, F.N., Han, S., Moskewicz, M.W., Ashraf, K., Dally, W.J., Keutzer, K.:
  Squeezenet: Alexnet-level accuracy with 50x fewer parameters and <0.5mb model
  size (2016)

\bibitem{jiang_eh-dnas_2021}
Jiang, Q., Zhang, X., Chen, D., Do, M.N., Yeh, R.A.: {EH}-{DNAS}:
  {End}-to-{End} {Hardware}-aware {Differentiable} {Neural} {Architecture}
  {Search}. arXiv:2111.12299 [cs]  (Nov 2021),
  \url{http://arxiv.org/abs/2111.12299}, arXiv: 2111.12299

\bibitem{jin2019rc}
Jin, X., Wang, J., Slocum, J., Yang, M.H., Dai, S., Yan, S., Feng, J.:
  Rc-darts: Resource constrained differentiable architecture search. arXiv
  preprint arXiv:1912.12814  (2019)

\bibitem{refbib2}
Jozefowicz, R., Zaremba, W., Sutskever, I.: An empirical exploration of
  recurrent network architectures. In: Proceedings of the 32nd International
  Conference on International Conference on Machine Learning - Volume 37. p.
  2342–2350. ICML'15, JMLR.org (2015)

\bibitem{kim2020puzzle}
Kim, J.H., Choo, W., Song, H.O.: Puzzle mix: Exploiting saliency and local
  statistics for optimal mixup (2020)

\bibitem{li2020sgas}
Li, G., Qian, G., Delgadillo, I.C., Müller, M., Thabet, A., Ghanem, B.: Sgas:
  Sequential greedy architecture search (2020)

\bibitem{lin2020mcunet}
Lin, J., Chen, W.M., Lin, Y., Gan, C., Han, S., et~al.: Mcunet: Tiny deep
  learning on iot devices. Advances in Neural Information Processing Systems
  \textbf{33},  11711--11722 (2020)

\bibitem{liu2018darts}
Liu, H., Simonyan, K., Yang, Y.: Darts: Differentiable architecture search.
  arXiv preprint arXiv:1806.09055  (2018)

\bibitem{liu2017learning}
Liu, Z., Li, J., Shen, Z., Huang, G., Yan, S., Zhang, C.: Learning efficient
  convolutional networks through network slimming (2017)

\bibitem{lyu_resource-constrained_2022}
Lyu, B., Yuan, H., Lu, L., Zhang, Y.: Resource-{Constrained} {Neural}
  {Architecture} {Search} on {Edge} {Devices}. IEEE Transactions on Network
  Science and Engineering  \textbf{9}(1),  134--142 (Jan 2022).
  \doi{10.1109/TNSE.2021.3054583}, conference Name: IEEE Transactions on
  Network Science and Engineering

\bibitem{ma2018shufflenet}
Ma, N., Zhang, X., Zheng, H.T., Sun, J.: Shufflenet v2: Practical guidelines
  for efficient cnn architecture design. In: Proceedings of the European
  conference on computer vision (ECCV). pp. 116--131 (2018)

\bibitem{munoz_enabling_2021}
Muñoz, J.P., Lyalyushkin, N., Akhauri, Y., Senina, A., Kozlov, A., Jain, N.:
  Enabling {NAS} with {Automated} {Super}-{Network} {Generation} (Dec 2021),
  \url{http://arxiv.org/abs/2112.10878}, arXiv:2112.10878 [cs]

\bibitem{Pham2018EfficientNA}
Pham, H., Guan, M.Y., Zoph, B., Le, Q.V., Dean, J.: Efficient neural
  architecture search via parameter sharing. In: ICML (2018)

\bibitem{real2019regularized}
Real, E., Aggarwal, A., Huang, Y., Le, Q.V.: Regularized evolution for image
  classifier architecture search (2019)

\bibitem{refbib1}
Real, E., Moore, S., Selle, A., Saxena, S., Suematsu, Y.L., Tan, J., Le, Q.V.,
  Kurakin, A.: Large-scale evolution of image classifiers. In: Proceedings of
  the 34th International Conference on Machine Learning - Volume 70. p.
  2902–2911. ICML'17, JMLR.org (2017)

\bibitem{sandler2018mobilenetv2}
Sandler, M., Howard, A., Zhu, M., Zhmoginov, A., Chen, L.C.: Mobilenetv2:
  Inverted residuals and linear bottlenecks. In: Proceedings of the IEEE
  conference on computer vision and pattern recognition. pp. 4510--4520 (2018)

\bibitem{srivastava2021variational}
Srivastava, A., Dutta, O., Gupta, J., Agarwal, S., AP, P.: A variational
  information bottleneck based method to compress sequential networks for human
  action recognition. In: Proceedings of the IEEE/CVF Winter Conference on
  Applications of Computer Vision. pp. 2745--2754 (2021)

\bibitem{szegedy2015going}
Szegedy, C., Liu, W., Jia, Y., Sermanet, P., Reed, S., Anguelov, D., Erhan, D.,
  Vanhoucke, V., Rabinovich, A.: Going deeper with convolutions. In:
  Proceedings of the IEEE conference on computer vision and pattern
  recognition. pp.~1--9 (2015)

\bibitem{tan2019platform}
Tan, M., Chen, B., Pang, R., Vasudevan, V., Sandler, M., Howard, A., Le~QV, M.:
  platform-aware neural architecture search for mobile. 2019 ieee. In: CVF
  Conference on Computer Vision and Pattern Recognition (CVPR). pp. 2815--2823
  (2019)

\bibitem{tan2019efficientnet}
Tan, M., Le, Q.: Efficientnet: Rethinking model scaling for convolutional
  neural networks. In: International conference on machine learning. pp.
  6105--6114. PMLR (2019)

\bibitem{wang2021rethinking}
Wang, R., Cheng, M., Chen, X., Tang, X., Hsieh, C.J.: Rethinking architecture
  selection in differentiable nas. arXiv preprint arXiv:2108.04392  (2021)

\bibitem{abc}
Wistuba, M.: Deep learning architecture search by neuro-cell-based evolution
  with function-preserving mutations. In: Berlingerio, M., Bonchi, F.,
  G{\"a}rtner, T., Hurley, N., Ifrim, G. (eds.) Machine Learning and Knowledge
  Discovery in Databases. pp. 243--258. Springer International Publishing, Cham
  (2019)

\bibitem{wu2019fbnet}
Wu, B., Dai, X., Zhang, P., Wang, Y., Sun, F., Wu, Y., Tian, Y., Vajda, P.,
  Jia, Y., Keutzer, K.: Fbnet: Hardware-aware efficient convnet design via
  differentiable neural architecture search. In: Proceedings of the IEEE/CVF
  Conference on Computer Vision and Pattern Recognition. pp. 10734--10742
  (2019)

\bibitem{wu_compiler-aware_2022}
Wu, Y., Gong, Y., Zhao, P., Li, Y., Zhan, Z., Niu, W., Tang, H., Qin, M., Ren,
  B., Wang, Y.: Compiler-{Aware} {Neural} {Architecture} {Search} for
  {On}-{Mobile} {Real}-time {Super}-{Resolution} (Jul 2022),
  \url{http://arxiv.org/abs/2207.12577}, arXiv:2207.12577 [cs, eess]

\bibitem{xiao_shapley-nas_2022}
Xiao, H., Wang, Z., Zhu, Z., Zhou, J., Lu, J.: Shapley-{NAS}: {Discovering}
  {Operation} {Contribution} for {Neural} {Architecture} {Search} (Jun 2022),
  \url{http://arxiv.org/abs/2206.09811}, arXiv:2206.09811 [cs]

\bibitem{xie2018snas}
Xie, S., Zheng, H., Liu, C., Lin, L.: Snas: stochastic neural architecture
  search. In: International Conference on Learning Representations (2018)

\bibitem{xiong2019resource}
Xiong, Y., Mehta, R., Singh, V.: Resource constrained neural network
  architecture search: Will a submodularity assumption help? In: Proceedings of
  the IEEE/CVF International Conference on Computer Vision. pp. 1901--1910
  (2019)

\bibitem{xu2019pc}
Xu, Y., Xie, L., Zhang, X., Chen, X., Qi, G.J., Tian, Q., Xiong, H.: Pc-darts:
  Partial channel connections for memory-efficient architecture search. arXiv
  preprint arXiv:1907.05737  (2019)

\bibitem{yang2021towards}
Yang, Y., You, S., Li, H., Wang, F., Qian, C., Lin, Z.: Towards improving the
  consistency, efficiency, and flexibility of differentiable neural
  architecture search. In: Proceedings of the IEEE/CVF Conference on Computer
  Vision and Pattern Recognition. pp. 6667--6676 (2021)

\bibitem{zhang2018shufflenet}
Zhang, X., Zhou, X., Lin, M., Sun, J.: Shufflenet: An extremely efficient
  convolutional neural network for mobile devices. In: Proceedings of the IEEE
  conference on computer vision and pattern recognition. pp. 6848--6856 (2018)

\bibitem{zheng_neural_2022}
Zheng, X., Fei, X., Zhang, L., Wu, C., Chao, F., Liu, J., Zeng, W., Tian, Y.,
  Ji, R.: Neural {Architecture} {Search} with {Representation} {Mutual}
  {Information}. In: 2022 {IEEE}/{CVF} {Conference} on {Computer} {Vision} and
  {Pattern} {Recognition} ({CVPR}). pp. 11902--11911. IEEE, New Orleans, LA,
  USA (Jun 2022). \doi{10.1109/CVPR52688.2022.01161},
  \url{https://ieeexplore.ieee.org/document/9878903/}

\bibitem{zhou2019bayesnas}
Zhou, H., Yang, M., Wang, J., Pan, W.: Bayesnas: A bayesian approach for neural
  architecture search (2019)

\bibitem{zhu2017trained}
Zhu, C., Han, S., Mao, H., Dally, W.J.: Trained ternary quantization (2017)

\end{thebibliography}
\end{small}
\newpage
\appendix
\title{Appendix}
\author{}
\institute{}
\maketitle
\section{Deriving cell architectures }
\begin{figure}[t]
\centering
   \includegraphics[width=\linewidth]{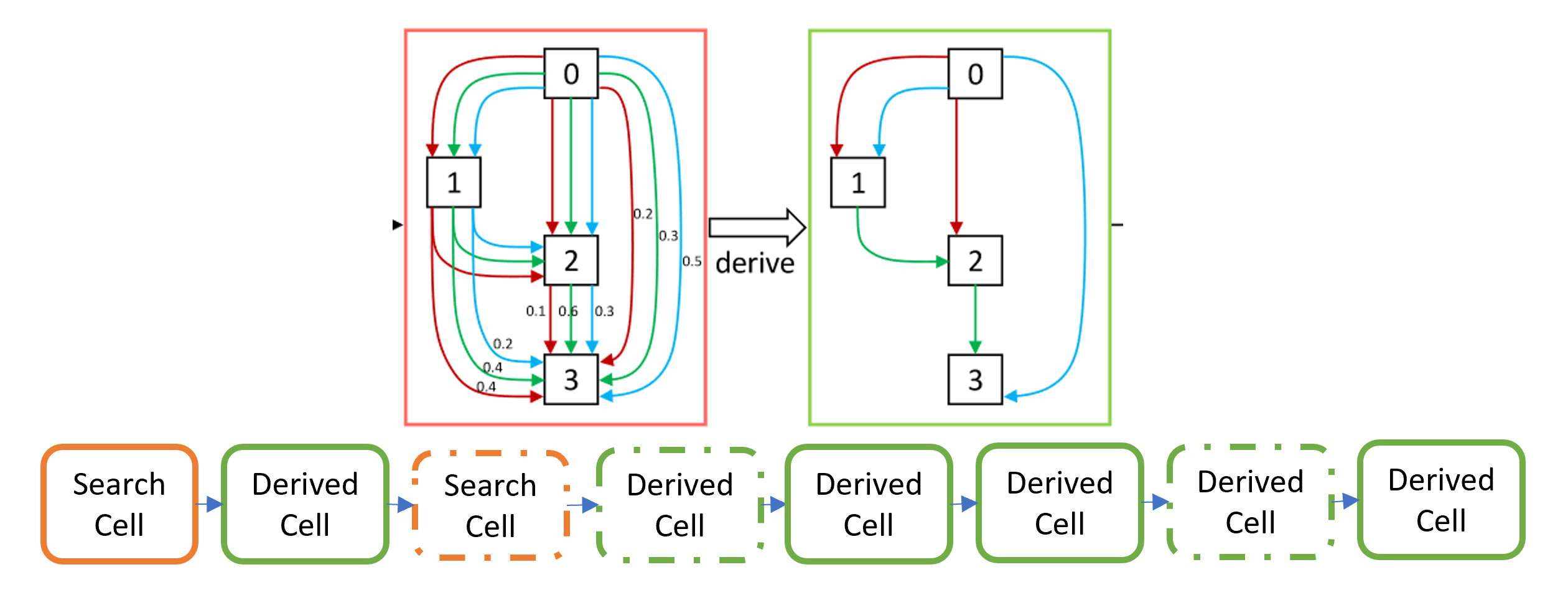}
    \caption{\small Top: shows the regular DARTS cell with nodes connected by weighted operations and the derived cell made of top-weighted operations. Bottom: Shows the network comprising the normal cell (bold border) and reduction cells (dotted border) with trainable architecture parameters (red border) and the derived cells (green border) without any architecture parameters.}
    \label{figtransit}
\vspace{-5mm}
\end{figure}

The searched cells are stacked to form the network whose weights are trained and evaluated. The layers of this network during the evaluation phase is varied from 4 to 20. It can be seen that the models searched with DARTS with only 2-cells perform equally well as those of 8-cell search for target model with layers more than 10. Hence, in our experiments, instead of training architecture parameters for all 8 cells, we train only 2 cells- one normal and the other reduction cell. The architecture of the other 6 cells stacked to form the network during search are derived from either the normal or the reduction cell as shown in Figure~\ref{figtransit}.
\section{Calculation of search-stage architecture size}
\begin{figure}[h!]
\centering
   \includegraphics[width=\linewidth]{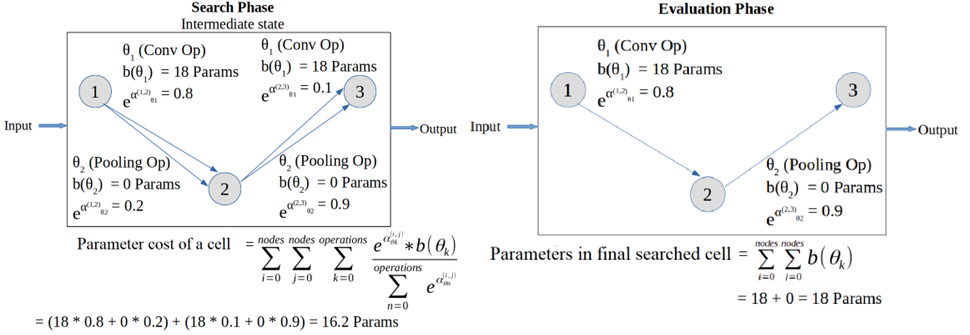}
    \caption{\small Demonstrates the calculation of memory size of a single cell in the architecture during - Left: search phase. Right: evaluation phase}
    \label{Fig-TacklingDifference}
\vspace{-5mm}
\end{figure}

The size of the architecture in the search phase $k_{s}$ is different from the architecture size in evaluation phase due to the softmax weighting factor in equation~\ref{eq4} (demonstrated in Figure ~\ref{Fig-TacklingDifference}). To address this, we introduce a tighter bound on the search constraint $K_{d^{\prime}}$, which is less than the device resource constraint $K_d$. A lookup graph (LUG) needs to be made for each dataset by varying $K_{d^\prime}$ within appropriate bounds and running the algorithm until convergence each time to obtain the corresponding device resource constraint $K_d$. The computation time of the LUG can be reduced by running the searches in parallel. 

\section{Algorithm}
The practical implementation of our resource-constrained gradient descent-based approach is illustrated in Algrorithm~\ref{alg:method}. 
\begin{algorithm}[t]
   \caption{ \small DCA-NAS - gradient descent based search method}
   \label{alg:method}
\begin{algorithmic}
\State Assign random weights to $\alpha^{i,j}$ on edges $i,j$ denoting weights of operations in the mixed set
\State Input look-up graph $G$ and device memory constraint $K_d$
\State Look-up corresponding search memory constraint $K_{d^{\prime}}$ from $G$
\State Calculate total search time memory size $k_s(\alpha)$ 
\While{$not$  $converged$}
    \State Calculate $\widetilde\mathcal{L}(w,\alpha,\lambda)=\mathcal{L}_{\text {val}}\left(w(\alpha), \alpha\right)
   +\lambda (k_{s}(\alpha)-K_{d^{\prime}})$
    \State Update weights $w$ by descending $\bigtriangledown_{w}\,\widetilde\mathcal{L}_{train}(w,\alpha,\lambda)$
    \State Update $\alpha$ by descending $\bigtriangledown_{\alpha}\,\widetilde\mathcal{L}_{val}(w^{*},\alpha,\lambda)$ 
    \State Calculate total search time memory size $k_s(\alpha)$
    \State Calculate loss as in equation~\ref{eq3a}
    \State Update $\lambda$ 
\EndWhile
\State Derive the final architecture based on the learned $\alpha$ by connecting the top weighted operations among the mixed set
\end{algorithmic}
\end{algorithm}

\section{Implementation Details}
The experiments with the smaller vision datasets-MNIST, FashionMNIST, CIFAR-10, Imagenet-16-120 and TinyImagenet were run on a single Tesla V100 GPU. Training and evaluation on Imagenet-1k was performed on a cluster containing eight V100 GPUs. \\
The super-net used for search with smaller vision datasets except Imagenet-1k consists of 8 cells, with 6 normal cells and 2 reduction cells, and an initial number of channels set to 16. Each cell has 6 nodes, with the first 2 nodes in cell k serving as input nodes. The super-net is trained for 50 epochs with a batchsize of 512, and optimized using SGD with a momentum of 0.9 and weight decay of $3e-4$. The learning rate is initially set to 0.2 and gradually reduced to zero using a cosine scheduler. Architecture parameters $\alpha$ are optimized using Adam optimizer, with a learning rate of $6e-4$, a momentum of $(0.5, 0.999)$, and a weight decay of $1e-3$. The search is run 5 times, and the architecture with the highest validation accuracy is chosen. For evaluation, the target-net has 20 cells, with 18 normal cells and 2 reduction cells, and an initial number of channels set to 36. The target-net is trained for 600 epochs with a batchsize of 96, optimized using SGD with a momentum of 0.9, weight decay of 3e-4, and gradient clipping of 5. The initial learning rate is set to 0.025 and gradually reduced to zero using a cosine scheduler. Additional settings include a cutout length of 16, dropout rate of 0.2, and use of an auxiliary head.
For Imagenet-1k, We reduce the input size from 224 × 224 to 28 × 28 using three convolution layers with a stride of 2. The super-net for search has 8 cells starting with 16 channels, and the target-net for evaluation has 14 cells starting with 48 channels. Both search and evaluation use a batch size of 1,024. In search, we train for 50 epochs with a learning rate of 0.5 (annealed down to zero using a cosine scheduler), and a learning rate of $6e-3$ for architecture parameters. In evaluation, we train for 250 epochs using the SGD optimizer with a momentum of 0.9 and a weight decay of $3e-5$, and adopt an auxiliary head and the label smoothing technique. \\
\section{Model performance by varying FLOPs constraint on CIFAR10, TinyImagenet and Imagenet-1k }
\begin{figure}[h!]
\begin{center}
\includegraphics[width=1\linewidth]{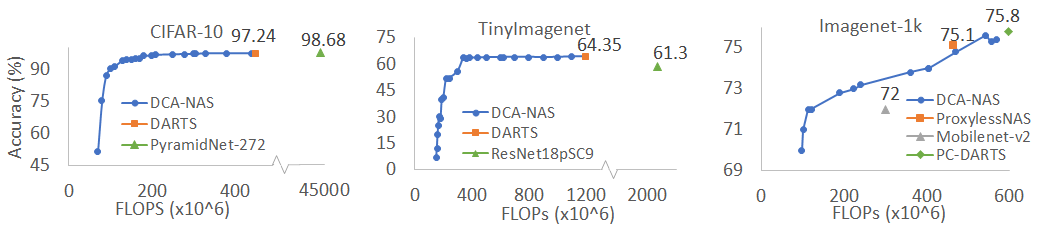}
\end{center}
    \caption{ Plots show that DCA-NAS method discovers models with fewer FLOPs than other NAS methods and manual architectures without sacrificing prediction performance. }
    \label{figflops}
\end{figure}

Instead of model parameters, we also experiment with FLOPs as the constraint in our objective function. As shown in Figure~\ref{figflops}, our method DCA-NAS retains performance till a certain FLOPs constraint, after which it degrades.  In comparison to manual architectures, our NAS approach yields models which require much smaller FLOPs and hence would have lower latency.
\vfill

\end{document}